\documentclass[11pt,a4paper]{article}
\usepackage[hyperref]{acl2021}
\usepackage{subcaption}
\usepackage{times}
\usepackage{latexsym}

\usepackage{graphicx}
\usepackage{microtype}
\usepackage{enumerate}
\usepackage{booktabs}
\usepackage{multirow}
\usepackage{makecell}
\usepackage{listings}
\usepackage[T1]{fontenc}
\usepackage{fixltx2e}

\aclfinalcopy 

\title{TransWiC at SemEval-2021 Task 2: Transformer-based Multilingual and Cross-lingual Word-in-Context Disambiguation}

\author{Hansi Hettiarachchi$^\heartsuit$, Tharindu Ranasinghe$^	\S$ \\

  $^\heartsuit$School of Computing and Digital Technology, Birmingham City University, UK \\
    
  $^\S$Research Group in Computational Linguistics, University of Wolverhampton, UK \\

  {\tt hansi.hettiarachchi@mail.bcu.ac.uk } \\
  {\tt tharindu.ranasinghe@wlv.ac.uk }} 

\date{}

\begin{document}
\maketitle

\begin{abstract}
Identifying whether a word carries the same meaning or different meaning in two contexts is an important research area in natural language processing which plays a significant role in many applications such as question answering, document summarisation, information retrieval and information extraction. Most of the previous work in this area rely on language-specific resources making it difficult to generalise across languages. Considering this limitation, our approach to SemEval-2021 Task 2 is based only on pretrained transformer models and does not use any language-specific processing and resources. Despite that, our best model achieves 0.90 accuracy for English-English subtask which is very compatible compared to the best result of the subtask; 0.93 accuracy. Our approach also achieves satisfactory results in other monolingual and cross-lingual language pairs as well. 
\end{abstract}

\section{Introduction}
Words' semantics have a dynamic nature which depends on the surrounding context \cite{pilehvar2019wic}. Therefore, the majority of words tends to be polysemous (i.e. have multiple senses). For few examples, words such as \textit{"cell"}, \textit{"bank"} and \textit{"report"} can be mentioned. Due to this nature in natural language, it is important to focus on word-in-context sense while extracting the meaning of a word which appeared in a text segment. Also, this is a critical requirement to many applications such as question answering, document summarisation, information retrieval and information extraction.  

Word Sense Disambiguation (WSD)-based approaches were widely used by previous research to tackle this problem \cite{loureiro2019liaad, Scarlini_Pasini_Navigli_2020}. WSD associates the word in a text with its correct meaning from a predefined sense inventory \cite{navigli2009word}. As such inventories, WordNet \cite{miller1995wordnet} and BabelNet \cite{NAVIGLI2012217} were commonly used. However, these approaches fail to generalise into different languages as these inventories are often limited to high resource languages. Targeting this gap, SemEval-2021 Task 2: Multilingual and Cross-lingual Word-in-Context Disambiguation is designed to capture the word sense without relying on fixed sense inventories in both monolingual and cross-lingual setting. In summary, this task is designed as a binary classification problem which predicts whether the target word has the same meaning or different meaning in different contexts of the same language (monolingual setting) or different languages (cross-lingual setting). 

This paper describes our submission to SemEval-2021 Task 2 \cite{martelli-etal-2021-mclwic}. Our approach is mainly focused on transformer-based models with different text pair classification architectures. We remodel the default text pair classification architecture and introduce several strategies that outperform the default text pair classification architecture for this task. For effortless generalisation across the languages, we do not use any language-specific processing and resources. In the subtasks where only a few training instances were available, we use few-shot learning and in the subtasks where there were no training instances were available, we use zero-shot learning taking advantage of the cross-lingual nature of the multilingual transformer models. 

The remainder of this paper is organised as follows. Section \ref{sec:related-work} describes the related work done in the field of word-in-context disambiguation. Details of the task data sets are provided in Section \ref{sec:data}. Section \ref{sec:transwic-architecture} describes the proposed architecture and Section \ref{sec:experimental-setup} provides the  experimental setup details. Following them, Section \ref{sec:results-and-evaluation} demonstrates the obtained results and Section \ref{sec:conclusions} concludes the paper with final remarks and future research directions.    

\section{Related Work} \label{sec:related-work}

\paragraph{Unsupervised systems} Majority of the unsupervised WSD systems use external knowledge bases like WordNet \cite{miller1995wordnet} and BabelNet \cite{NAVIGLI2012217}. For each input word, its correct meaning according to the context can be found using graph-based techniques from those external knowledge bases. However, these approaches are only limited to the languages supported by used knowledge bases. More recent works like \citet{hettiarachchi-ranasinghe-2020-brums, ranasinghe-etal-2019-enhancing} propose to use stacked word embeddings \cite{akbik-etal-2018-contextual} obtained by general purpose pretrained  contextualised word embedding models such as BERT \cite{devlin-etal-2019-bert} and Flair \cite{akbik-etal-2019-pooled} for unsupervised WSD. Despite their ability to scale over different languages, unsupervised approaches fall behind supervised systems in terms of accuracy.


\paragraph{Supervised systems}
Supervised systems rely on semantically-annotated corpora for training \cite{raganato-etal-2017-neural,bevilacqua-navigli-2019-quasi}. Early approaches were based on traditional machine learning algorithms like support vector machines \cite{iacobacci-etal-2016-embeddings}. With the word embedding-based approaches getting popular in natural language processing tasks, more recent approaches on WSD were based on neural network architectures \cite{melamud-etal-2016-context2vec, raganato-etal-2017-neural}. However, they rely on large manually-curated training data to train the machine learning models which in turn hinders the ability of these approaches to scale over unseen words and new languages. More recently, contextual representations of words have been used in WSD where the contextual representations have been employed for the creation of
sense embeddings \cite{peters-etal-2018-deep}. However, they also rely on sense-annotated corpora to gather contextual information for each sense, and hence are limited to languages for which gold annotations are available. A very recent approach SensEmBERT \cite{Scarlini_Pasini_Navigli_2020} provide WSD by leveraging the mapping between senses and Wikipedia pages, the relations among BabelNet synsets and the expressiveness of contextualised embeddings, getting rid of manual annotations. However, SensEmBERT \cite{Scarlini_Pasini_Navigli_2020} only supports five languages making it difficult to use with other languages.

Considering the limitations of the above methods, in this paper we propose an approach which is based on general purpose transformer models and does not rely on external knowledge bases. Also, our approach shows strong few-shot/zero-shot learning performance removing the hurdle of having manually-curated training data for each language pair.


\section{Data} \label{sec:data}
The data set used for SemEval-2021 Task 2 is designed targeting a binary classification problem following \citet{pilehvar2019wic}. To preserve the multilinguality and cross-linguality of the task, five different languages: English, Arabic, French, Russian and Chinese have been considered for data set preparation. In the monolingual setting, per instance, a sentence pair written in the same language is provided with a targeted lemma to predict whether it has the same meaning (True) or different meanings (False) in both sentences. In the cross-lingual setting, each sentence pair is written in two different languages with the same prediction requirement. Few samples from the monolingual and cross-lingual data sets are shown in Table \ref{tab:data-sample}. 


\renewcommand{\arraystretch}{1.2}
\begin{table*}[t]
\begin{center}
\small
\begin{tabular}{l c c c c}
\toprule
& \bf Lang. & \bf Sentence 1 & \bf Sentence 2 & \bf Label \\
\midrule
\multirow{2}{*}{\bf ML} & \makecell[l]{fr-fr} & \makecell[l]{la \color{red}souris \color{black} mange le fromage} & \makecell[l]{le chat court après la \color{red}souris \color{black}} & \makecell{T} \\
\cmidrule(r){2-5}
& \makecell[l]{en-en} & \makecell[l]{In the private \color{red} sector \color{black}, activities are guided by\\ the motive to earn money.} & \makecell[l]{The volume V of the \color{red}sector \color{black} is related to the \\ area A of the cap.} & \makecell{F}  \\
\midrule
\multirow{2}{*}{\bf CL} & \makecell[l]{en-fr} & \makecell[l]{click the right \color{red}mouse \color{black} button} & \makecell[l]{le chat court après la \color{red}souris \color{black}} & \makecell{F} \\
\cmidrule(r){2-5}
& \makecell[l]{en-fr} & \makecell[l]{Any alterations which it is proposed to make\\ as a result of this review are to be \color{red}reported \color{black}\\ to the Interdepartmental Committee on\\ Charter Repertory for its approval.} & \makecell[l]{Il a aussi été \color{red}indiqué \color{black} que, selon les dossiers\\ médicaux, Justiniano Hurtado Torre était\\ mort de maladie.} & \makecell{T} \\
\bottomrule
\end{tabular}
\end{center}
\caption{Monolingual (ML) and cross-lingual (CL) sentence pair samples with targeted lemma (highlighted in red colour) and label (T:True, F:False). Lang. column represent the languages which are indicated using ISO 639-1 codes\footnote{ISO 639-1 language codes are available on\url{https://www.loc.gov/standards/iso639-2/php/code_list.php}}} 
\label{tab:data-sample}
\end{table*}

The monolingual data set covers the language pairs: en-en, ar-ar, fr-fr, ru-ru and zh-zh. For each language, 8-instance trial data sets with labels were provided to give an insight into the task. As training data, 8,000 labelled instances were provided only for the English language and as dev data, 1,000 labelled instances were provided per each language. To use with final evaluation, for each language, 1,000-instance test data sets were provided.

The cross-lingual data set covers the language pairs: en-ar, en-fr, en-ru and en-zh. Similar to the monolingual data set, 8-instance trial data sets with labels were provided for each language pair. However, no training or dev data sets were provided for the cross-lingual setting. To use with the final evaluation, 1,000-instance test data sets were provided per each language pair.

\begin{figure*}[ht]
\centering
\includegraphics[scale=0.4]{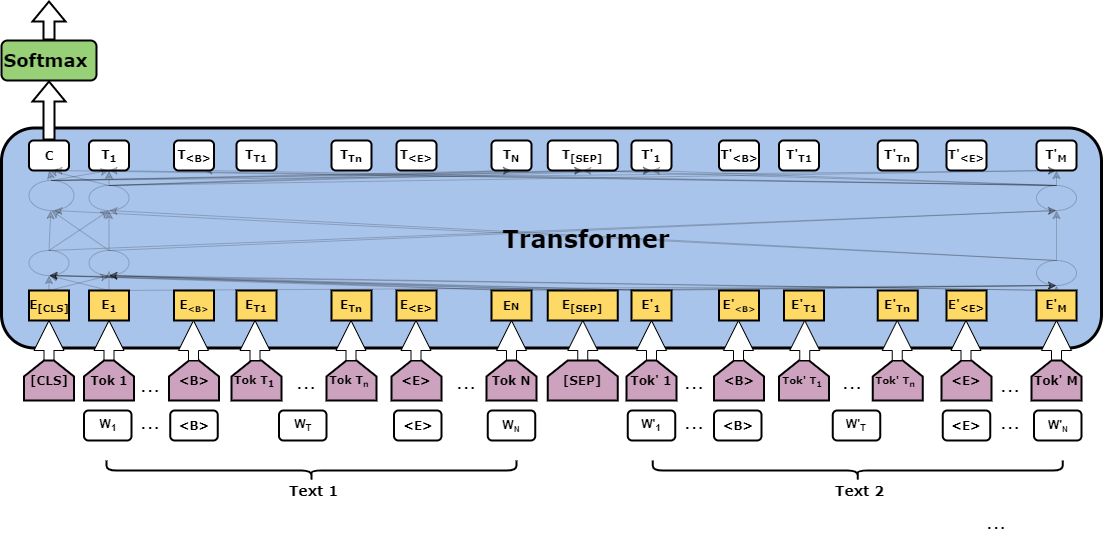}
\caption{Default sentence pair classification architecture - ([CLS] Strategy).W\textsubscript{T} is the target word.}
\label{fig:architecture}
\end{figure*}

\begin{figure*}

  \begin{subfigure}[b]{8cm}
    \centering\includegraphics[scale=0.19]{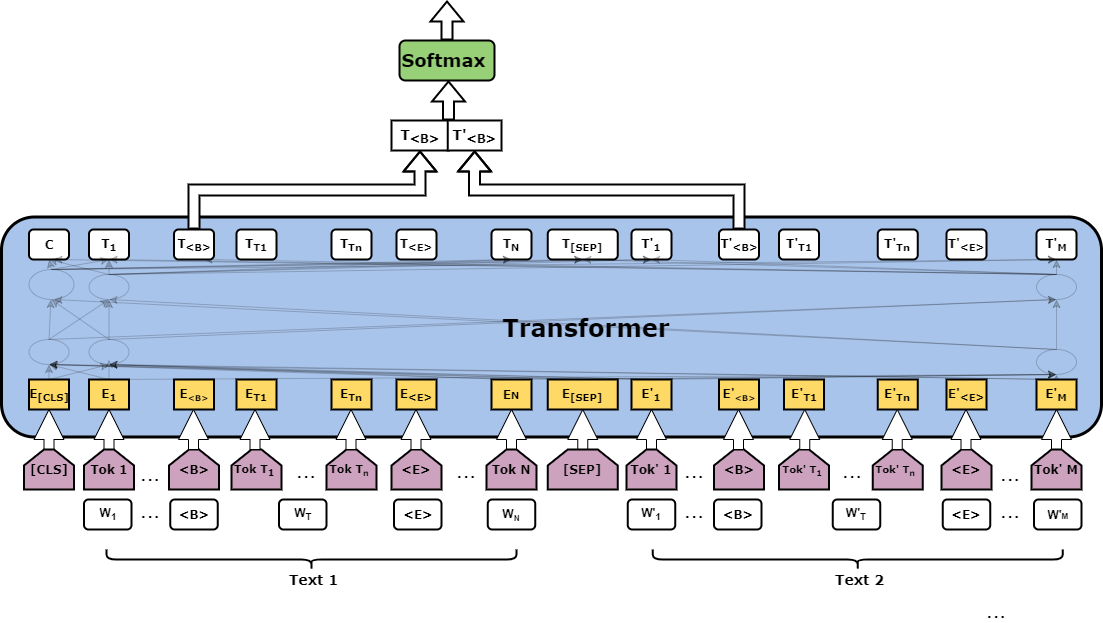}
    \caption{<B> Strategy}
    \label{fig:b_strategy}
  \end{subfigure}
  \begin{subfigure}[b]{8cm}
    \centering\includegraphics[scale=0.19]{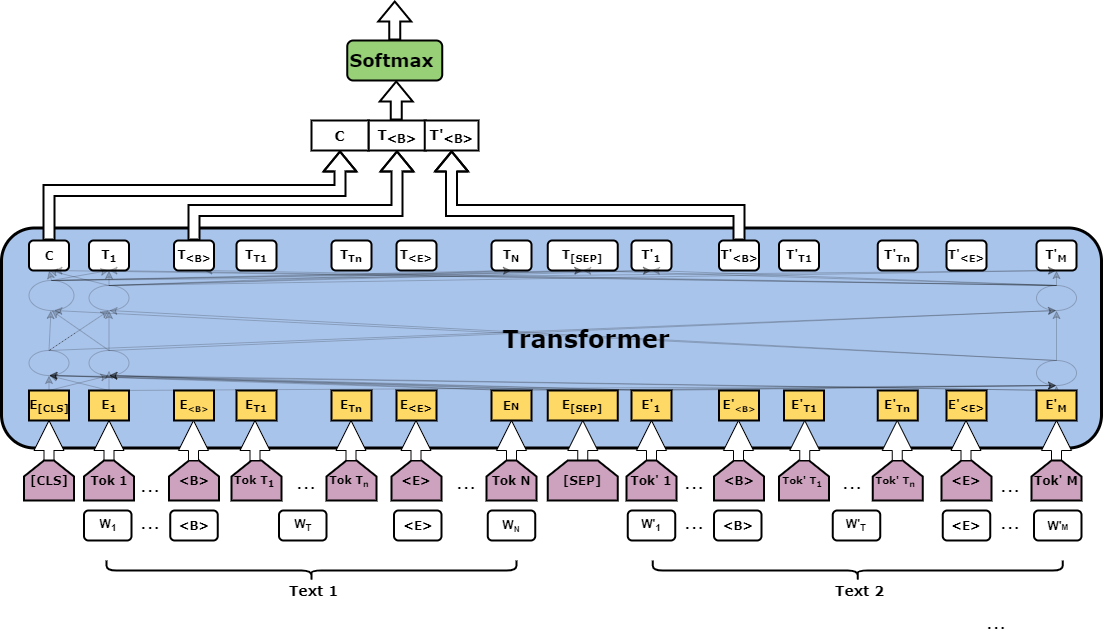}
    \caption{<B> + [CLS] Strategy}
    \label{fig:b_cls_strategy}
  \end{subfigure}
   \begin{subfigure}[b]{8cm}
    \centering\includegraphics[scale=0.19]{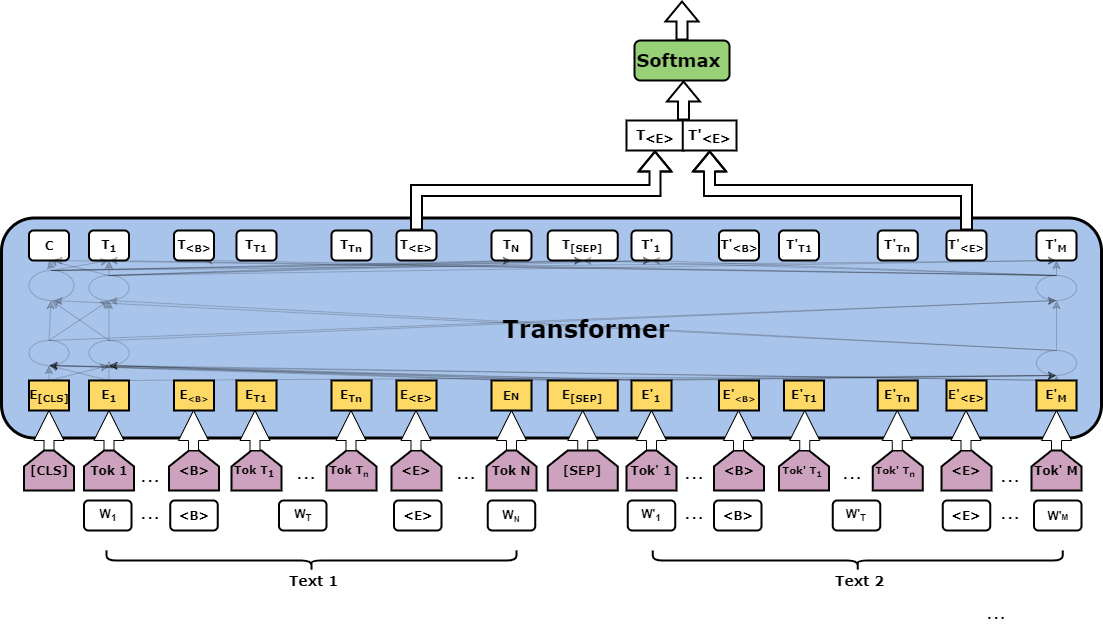}
    \caption{<E> Strategy}
    \label{fig:e_strategy}
  \end{subfigure}
  \begin{subfigure}[b]{8cm}
    \centering\includegraphics[scale=0.19]{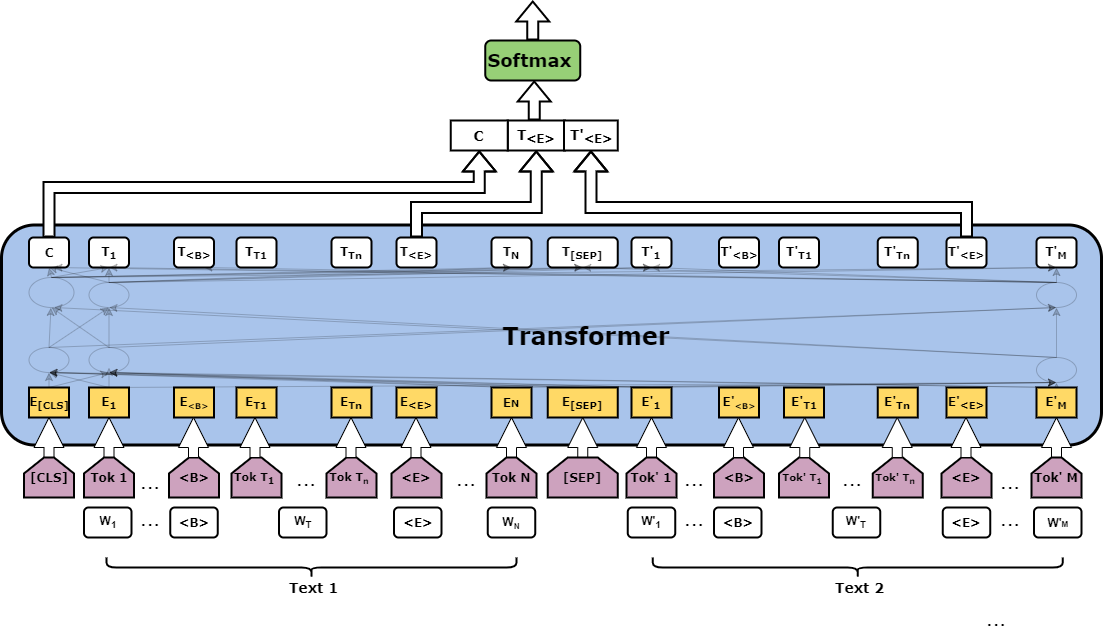}
    \caption{<E> + [CLS] Strategy}
    \label{fig:e_cls_strategy}
  \end{subfigure}
   \begin{subfigure}[b]{8cm}
    \centering\includegraphics[scale=0.19]{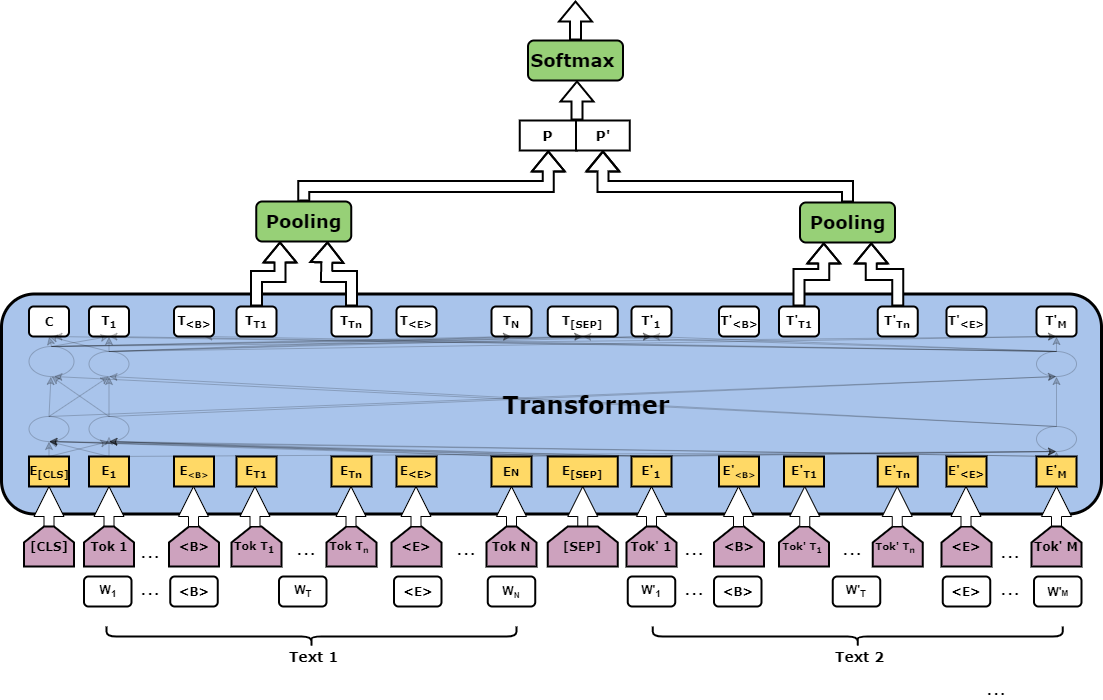}
    \caption{Entity Pool Strategy}
    \label{fig:entity_pool}
  \end{subfigure}
  \begin{subfigure}[b]{8cm}
    \centering\includegraphics[scale=0.19]{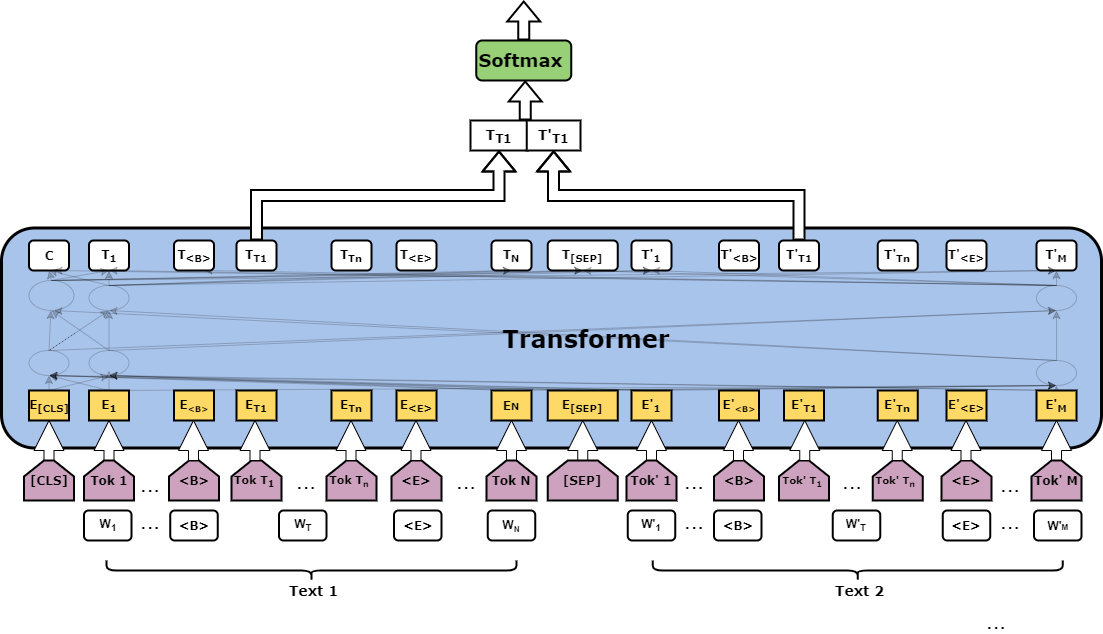}
    \caption{Entity First Strategy}
    \label{fig:entity_first}
  \end{subfigure}
   \begin{subfigure}[b]{8cm}
    \centering\includegraphics[scale=0.19]{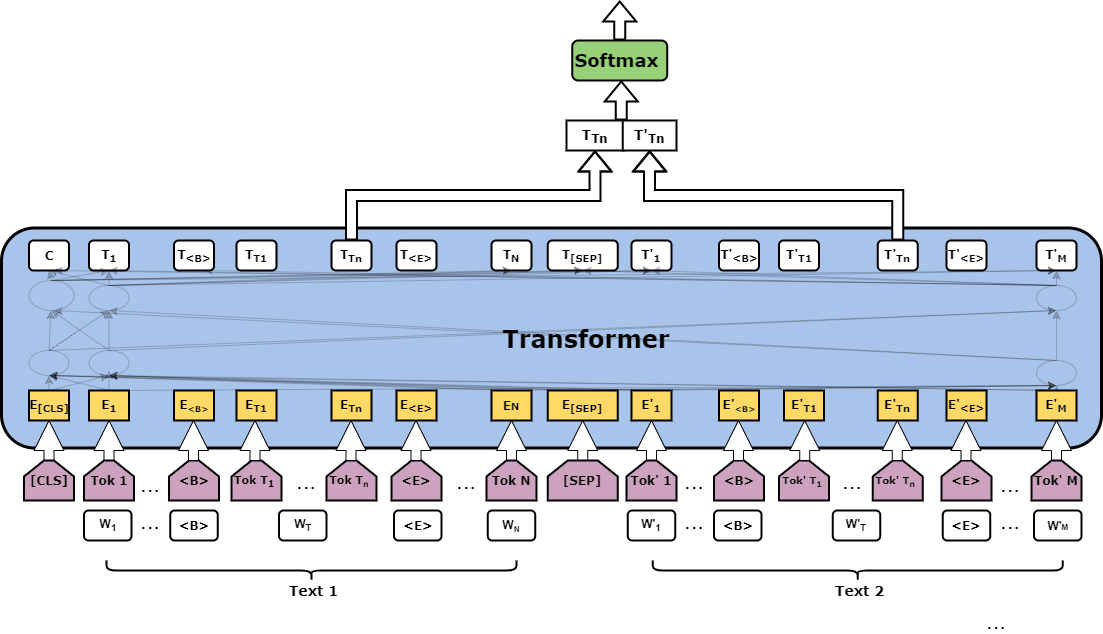}
    \caption{Entity Last Strategy}
    \label{fig:entity_last}
  \end{subfigure}
  \begin{subfigure}[b]{8cm}
    \centering\includegraphics[scale=0.19]{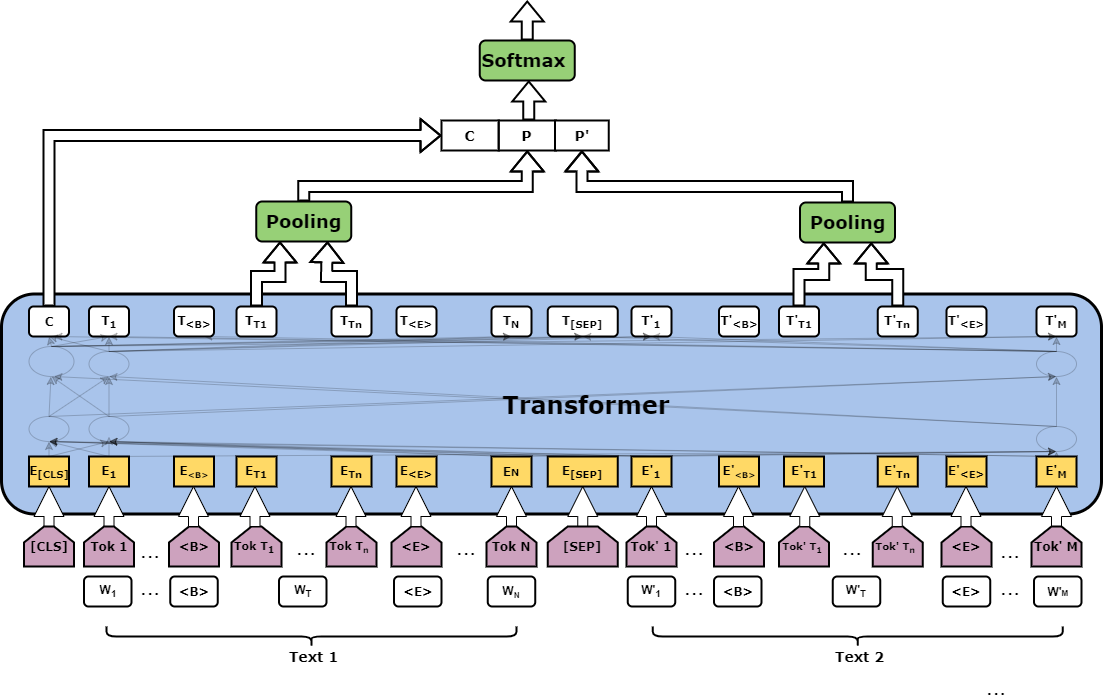}
    \caption{[CLS] + Entity Pool Strategy}
    \label{fig:cls_entity_pool}
  \end{subfigure}
   \begin{subfigure}[b]{8cm}
    \centering\includegraphics[scale=0.19]{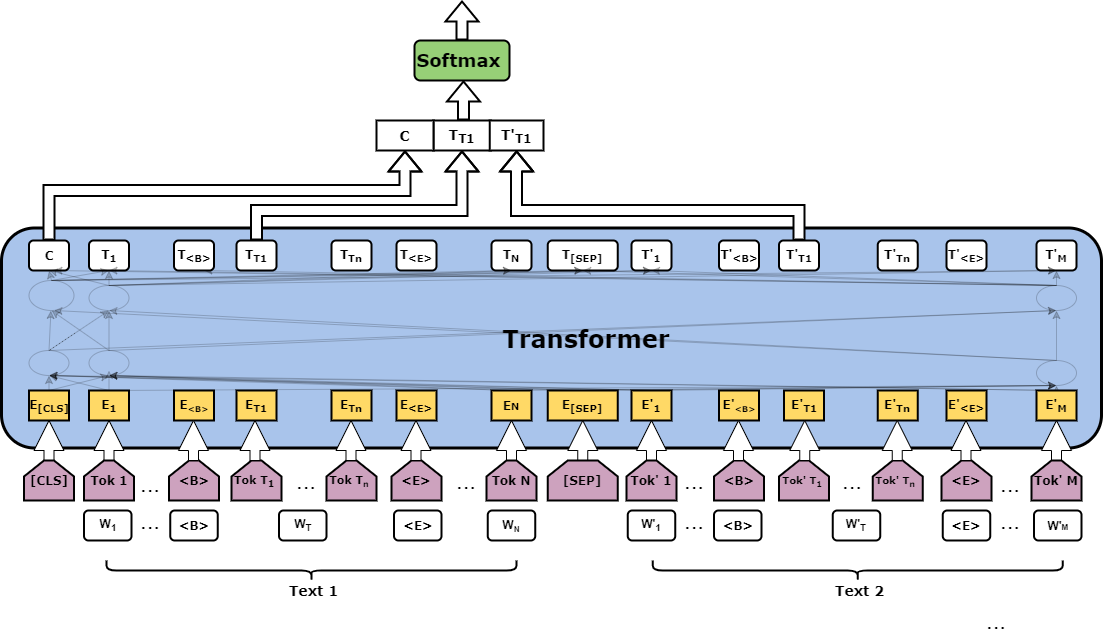}
    \caption{[CLS] + Entity First Strategy}
    \label{fig:cls_entity_first}
  \end{subfigure}
  \begin{subfigure}[b]{8cm}
    \centering\includegraphics[scale=0.19]{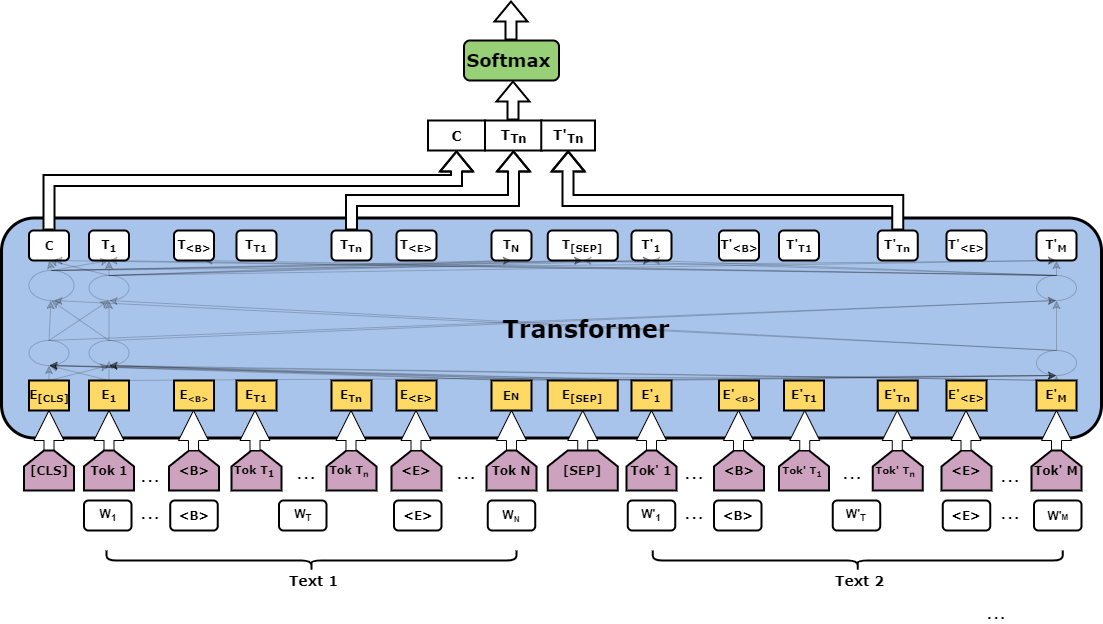}
    \caption{[CLS] + Entity Last Strategy}
    \label{fig:cls_entity_last}
  \end{subfigure}
\caption{Strategies in the TransWiC Framework. W\textsubscript{T} is the target word.}
\label{fig:architectures}
\end{figure*}

\section{TransWiC Architecture} \label{sec:transwic-architecture}
The main motivation behind the TransWiC architecture is the success transformer-based architectures had in various natural language processing tasks like offensive language identification \cite{ranasinghe-hettiarachchi-2020-brums, ranasinghe2019brums, pitenis-etal-2020-offensive}, offensive spans identification \cite{mudes, ranasinghe2021semeval}, language detection \cite{jauhiainen2021} question answering \cite{yang-etal-2019-end-end-open} etc. Apart from providing strong results compared to RNN based architectures \cite{hettiarachchi-ranasinghe-2019-emoji, ranasinghe2019brums}, transformer models like BERT \cite{devlin-etal-2019-bert}, XLM-R \cite{conneau-etal-2020-unsupervised} provide pretrained language models that support more than 100 languages. This is a huge benefit when compared to the models like SensEmBERT \cite{Scarlini_Pasini_Navigli_2020} which supports only five languages. Furthermore, multilingual and cross-lingual models like multilingual BERT \cite{devlin-etal-2019-bert} and XLM-R \cite{conneau-etal-2020-unsupervised} have shown strong transfer learning performance across scarce-resourced languages which can be useful in non-English monolingual subtasks where there are fewer training examples and cross-lingual subtasks where there are no training examples available \cite{ranasinghe-zampieri-2020-multilingual, ranasinghe2021tallip, ranasinghe2020wlv}. Therefore we took the general purpose transformers like BERT \cite{devlin-etal-2019-bert} and XLM-R \cite{conneau-etal-2020-unsupervised}, reworked their sentence pair classification architecture with so called \textit{strategies} described below to perform well in word-in-context disambiguation task. 

\paragraph{Preprocessing} As a preprocessing step we add two tokens to the transformer model's vocabulary: \textsc{<B>} and \textsc{<E>}. We place them around the target word in both sentences. For example, the sentence \textit{"la souris mange le fromage"} with the target word \textit{"souris"} will be changed to \textit{"la <B> souris <E> mange le fromage"}.


\begin{enumerate}[i]
    \item \textbf{[CLS] Strategy} - This is the default sentence pair classification architecture with transformers \cite{devlin-etal-2019-bert} where the two sentences are concatenated with a [SEP] token and passed through a transformer model. Then the output of the [CLS] token is fed into a softmax layer to predict the labels (Figure \ref{fig:architecture}).
    
    \item \textbf{<B> Strategy} - We concatenate the output of two <B> tokens of the two sentences and feed it into a softmax layer to predict the labels (Figure \ref{fig:b_strategy}). 
    
    \item \textbf{<B> + [CLS] Strategy} - We concatenate the output of two <B> tokens of the two sentences with the [CLS] token and feed it into a softmax layer to predict the labels (Figure \ref{fig:b_cls_strategy}). 
    
    \item \textbf{<E> Strategy} - Output of the two <E> tokens of the two sentences are concatenated and feed into a softmax layer to predict the labels (Figure \ref{fig:e_strategy}).
    
    \item \textbf{<E> + [CLS] Strategy} - We concatenate the output of two <E> tokens of the two sentences with the [CLS] token and feed it into a softmax layer to predict the labels (Figure \ref{fig:e_cls_strategy}).

    \item \textbf{Entity Pool Strategy} - To effectively deal with rare words, transformer models use sub-word units or WordPiece tokens as the input to build the models \cite{devlin-etal-2019-bert}. Therefore, there is a possibility that one target word can be separated into several sub-words. In this strategy, we generate separate fixed-length embeddings for each target word by passing its sub-word outputs through a pooling layer. The pooled outputs are concatenated and fed into a softmax layer to predict the labels (Figure \ref{fig:entity_pool}). 
    
    
    \item \textbf{Entity First Strategy} - Similar to the previous strategy, instead of using all the sub-words of the target word, we only use the output of the first sub-word in this strategy. We feed the concatenation of these outputs into a softmax layer to predict the labels (Figure \ref{fig:entity_first}). 
    
    \item \textbf{Entity Last Strategy} - Similar to the \textit{Entity First Strategy} instead of the first sub-word, we use the last sub-word to represent the target word. We feed their concatenation into a softmax layer to predict the labels (Figure \ref{fig:entity_last}). 
    
    \item \textbf{[CLS] + Entity Pool Strategy} - We concatenate the pooled outputs generated by \textit{Entity Pool Strategy} with the [CLS] token and feed it into a softmax layer to predict the labels (Figure \ref{fig:cls_entity_pool}). 
    
    \item \textbf{[CLS] + Entity First Strategy} - Similar to the \textit{[CLS] + Entity Pool Strategy}, instead of the pooled outputs, we concatenate the first sub-word output of the target words with [CLS] token and feed it into a softmax layer to predict the labels (Figure \ref{fig:cls_entity_first}). 
    
     \item \textbf{[CLS] + Entity Last Strategy} - In this strategy, we concatenate the last sub-word output of the target words with [CLS] token and feed it into a softmax layer to predict the labels (Figure \ref{fig:cls_entity_last}).

\end{enumerate}

\section{Experimental Setup} \label{sec:experimental-setup}
This section describes the training data and hyperparameter configurations used during the experiments. 

\subsection{Training Configurations}
\paragraph{English-English}  For the English-English subtask, we performed training on the English-English training data for each strategy mentioned above. During the training process, the parameters of the transformer model, as well as the parameters of the subsequent layers, were updated. We used the saved model from a particular strategy to get predictions for the English-English test set for that particular strategy.

\paragraph{Other Monolingual} Since there were less training data available for non-English monolingual datasets, we followed a \textit{few-shot} learning approach mentioned in \citet{ranasinghe-etal-2020-transquest, ranasinghe-etal-2020-transquest-wmt2020}. When we are starting the training for non-English monolingual language pairs, rather than training a model from scratch, we initialised the weights saved from the English-English experiment. Then we performed training on the dev data for each language pair separately. Similar to English-English experiments, during the training process, the parameters of the transformer model, as well as the parameters of the subsequent layers, were updated. 

\paragraph{Crosslingual} Since there were no training data available for cross-lingual datasets, we followed a \textit{zero-shot} approach for them. Multilingual and cross-lingual transformer models like multilingual BERT and XLM-R show strong cross-lingual transfer learning performance. They can be trained on one language; typically a resource-rich language and can be used to perform inference on another language. The cross-lingual nature of the transformer models has provided the ability to do this \cite{ranasinghe-etal-2020-transquest}. Therefore, we used the models trained on the English-English dataset to get predictions for cross-lingual datasets.

\subsection{Hyperparameter Configurations}
We used a Nvidia Tesla K80 GPU to train the models. We divided the input dataset into a training set and a validation set using 0.8:0.2 split. We predominantly fine-tuned the learning rate and the number of epochs of the classification model manually to obtain the best results for the validation set. We obtained $1e^-5$ as the best value for the learning rate and 3 as the best value for the number of epochs. We performed \textit{early stopping} if the validation loss did not improve over 10 evaluation steps. The rest of the hyperparameters which we kept as constants are mentioned in the Appendix. When performing training, we trained five models with different random seeds and considered the majority-class self ensemble mentioned in \citet{hettiarachchi-ranasinghe-2020-infominer} to get the final predictions.


\section{Results and Evaluation} \label{sec:results-and-evaluation}
Organisers used the accuracy as the evaluation metric as shown in Equation \ref{equ:acc} where TP is True Positive, TN is True Negative, FP is False Positive and FN is False Negative. 

\begin{equation}
 Accuracy = \frac{TP+TN}{TP+TN+FP+FN}   
 \label{equ:acc}
\end{equation}

Since there were less or no training data available for other monolingual and cross-lingual settings, we trained and evaluated models for each of our strategies using English-English training and dev sets. Then the best models are picked to use with few-shot and zero-shot learning approaches. We report the results obtained by English-English evaluation in Table \ref{tab:dev}. In the BERT column, we report the results of the bert-large-cased model while in the XLM-R column, we report the results of the xlm-r-large model.



\renewcommand{\arraystretch}{1.2}
\begin{table}[!ht]
\centering
\scalebox{1.0}{
\small
\begin{tabular}{lcc}
\toprule
\bf Strategy &   \bf BERT & \bf XLM-R \\ \hline

CLS &  0.8350   & 0.7860     \\
<B> &  0.8450   & 0.8750$^{\ddag}$     \\
<B> + CLS & 0.8590   & \textbf{0.8810$^{\ddag}$}     \\
<E> &  0.6672   & 0.5590     \\
<E> + CLS &   0.6982  & 0.5630     \\
Entity Pool &  0.8420   &  0.8521    \\
Entity First &  0.8390   &  0.8462    \\
Entity Last &  0.8550   &  0.8660   \\
CLS + Entity Pool &  0.8570   &  0.8700$^{\ddag}$    \\
CLS + Entity First &  0.8540   & 0.8580     \\
CLS + Entity Last &  0.8568   &  0.8610    \\
\bottomrule
\end{tabular}
}
\caption{TransWiC accuracy in English-English dev set for each strategy. Best is in Bold. Submitted systems are marked with \ddag}
\label{tab:dev}
\end{table}

\renewcommand{\arraystretch}{1.2}
\begin{table*}[t]
\begin{center}
\small
\begin{tabular}{l l  c c c c c c c c c} 
\toprule
& & \multicolumn{5}{c}{\bf Monolingual} & \multicolumn{4}{c}{\bf Crosslingual}\\\cmidrule(r){3-7}\cmidrule(r){8-11}
 &{\bf Strategy} & en-en & ar-ar
 & fr-fr & ru-ru & zh-zh & en-ar  & en-fr  & en-ru & en-zh \\
\midrule
\multirow{3}{*}{\bf I} & <B> + [CLS] & \textbf{0.9040}
 & 0.7800 & \textbf{0.7970} & \textbf{0.7610} & \textbf{0.6210} & \textbf{0.6690} & \textbf{0.5860} & \textbf{0.6900} & \textbf{0.7640} \\

& <B> & 0.8980 & \textbf{0.7980} & 0.7760 & 0.7160 & 0.6090 & 0.6260 & 0.5850 & 0.6770 & 0.7280 \\
& [CLS] + Entity Pool & 0.8400 & 0.7621 & 0.7321 & 0.6954 & 0.5880 & 0.5921 & 0.5572 & 0.6561 & 0.7002 \\
\midrule
\multirow{1}{*}{\bf II} & Best System & 0.9330 & 0.8480 & 0.8750 & 0.8740 & 0.9100 & 0.8910 & 0.8910 & 0.8940 & 0.9120 \\
\bottomrule
\end{tabular}
\end{center}
\caption{Row I shows the accuracy scores for the test set with strategies submitted. Best results for each language pair with our strategies are in bold. Row II shows the accuracy scores for the test set with the best system submitted for each language pair.} 
\label{tab:test}
\end{table*}

As shown in Table  \ref{tab:dev}, some strategies outperformed the default sentence pair classification architecture. Among all experimented strategies <B> + [CLS] strategy performed best. Usually, multilingual transformer models like XLM-R do not outperform the language-specific transformer models. Surprisingly, in this task XLM-R models outperform bert-large models. We selected three best performing models for the submission; XLM-R <B> + [CLS], XLM-R <B> and XLM-R [CLS] + Entity Pool. 

Since multilingual models provided the best results for the English-English dataset, it provided an additional advantage as they can be used directly in other language pairs too as mentioned in Section \ref{sec:experimental-setup}. For other language pairs, we did not perform any evaluation due to the lack of data availability. We trusted the cross-lingual performance of XLM-R and used the best three models of the English-English experiment. For the rest of the monolingual pairs, we used the few-shot learning approach using the given dev sets and for the cross-lingual pairs, we used the zero-shot learning approach mentioned in Section \ref{sec:experimental-setup}.


We report the results we got for the test set in Table \ref{tab:test}. According to the results, <B> + [CLS] strategy performs best in all the language pairs except Ar-Ar, where <B> strategy outperforms <B> + [CLS] strategy. When compared to the best models submitted to each language pair, our approach shows very competitive results in the majority of the monolingual language pairs. However, we believe that the cross-lingual performance of our methodology should be improved. Nonetheless, we believe that as a methodology that did not use any language-specific resources and did not see any language-specific data, the results are at a satisfactory level.

\section{Conclusions} \label{sec:conclusions}
In this paper, we presented our approach for tackling the SemEval-2021 Task 2: Multilingual and Cross-lingual Word-in-Context Disambiguation. We use the pretrained transformer models and remodel the sentence pair classification architecture for this task with several strategies. Our best strategies outperform the default sentence pair classification setting for English-English. For other monolingual language pairs, we use the few-shot learning approach while for cross-lingual language pairs we use the zero-shot approach. Our results are compatible with the best systems submitted for each language pair and are at a satisfactory level given the fact that we did not use any language-specific processing nor resources. 

As future work, we would be looking to improve our results more with new strategies. We would like to experiment with whether adding language-specific processing and resources would improve the results. We are keen to add different neural network architectures like Siamese transformer networks \cite{reimers-gurevych-2019-sentence} that perform well in sentence pair classification tasks \cite{ranasinghe-etal-2019-semantic, malstm} to the TransWiC framework. Furthermore, we are hoping to work in a multi-task environment and experiment whether transfer learning from a similar task like semantic textual similarity \cite{cer-etal-2017-semeval} would improve the results for this task.

\bibliographystyle{acl_natbib}
\bibliography{acl2021}

\appendix
\section{Appendix}

A summary of hyperparameters and their values used to obtain the reported results are mentioned in Table \ref{tab:params}. The optimised hyperparameters are marked with $\ddag$ and their optimal values are reported. The rest of the hyperparameter values are kept as constants. 

\renewcommand{\arraystretch}{1.2}
\begin{table}[!ht]
\centering
\scalebox{1.0}{
\small
\begin{tabular}{ll}
\toprule
\bf Parameter &   \bf Value \\ \hline
learning rate$^{\ddag}$ & $1e^{-5}$\\
number of epochs$^{\ddag}$ & $3$\\
adam epsilon & $1e^{-8}$\\
warmup ration & 0.1\\
warmup steps & 0\\
max grad norm & 1.0\\
max seq. length & 120\\
gradient accumulation steps & 1\\
\bottomrule
\end{tabular}
}
\caption{Hyperparameter specifications}
\label{tab:params}
\end{table}

\end{document}